\documentclass[11pt]{article}

\usepackage[final]{acl}

\usepackage{times}
\usepackage{latexsym}

\usepackage[T1]{fontenc}
\usepackage[utf8]{inputenc}

\usepackage{microtype}

\usepackage{inconsolata}

\usepackage[table]{xcolor}
\usepackage{graphicx}

\usepackage{amsmath, amssymb}
\usepackage{algorithm}
\usepackage{algorithmic}
\usepackage{multirow}
\usepackage{booktabs}
\usepackage{tcolorbox}
\tcbuselibrary{breakable, skins}
\usepackage{caption}

\title{Valid $\neq$ Necessary: Diagnosing Latent Inefficiency in Chain-of-Thought}

\author{
  Daeyeop Lee$^{1,2}$ \and Hwanjo Yu$^{2}$\thanks{\ \ Corresponding author.} \\
  $^1$KT Corporation \\
  $^2$Pohang University of Science and Technology \\
  \texttt{daeyeop.lee@kt.com} \\
  \texttt{hwanjoyu@postech.ac.kr}
}

\newcommand{\BENCHMARK}{RIV-GSM8K}  % 벤치마크 이름
\newcommand{\METRIC}{CAID}          % 제안하는 메트릭 이름
\newcommand{\FRAMEWORK}{PACE}       % 제안하는 프레임워크 이름

\begin{document}
\maketitle
\begin{abstract}
Chain-of-Thought (CoT) prompting has significantly advanced the reasoning capabilities of Large Language Models (LLMs), yet it often incurs substantial computational costs due to ``over-reasoning''---the generation of redundant, verbose, or irrelevant steps. While existing reasoning step evaluators effectively detect logical fallacies and factual errors, our analysis reveals a critical blind spot: they fail to penalize ``valid but inefficient'' reasoning steps that inflate token usage without contributing to the solution. To systematically diagnose this limitation, we introduce \textbf{\BENCHMARK}, a diagnostic benchmark injected with five distinct types of inefficiencies, including circular reasoning and excessive decomposition. Diagnostic experiments reveal that state-of-the-art evaluators struggle to distinguish these inefficiencies from necessary reasoning. To address this gap, we propose \textbf{\METRIC} (Context-Aware Information Density), a training-free metric grounded in information theory that identifies low-utility steps. To validate the metric's practical utility, we apply it within \textbf{\FRAMEWORK}, a post-hoc compression strategy. Additional control experiments show that the gains of \FRAMEWORK{} are not explained by trivial pruning: compared with random step removal and PRM-based compression baselines, it preserves accuracy at substantially higher compression rates. Empirical results on GSM8K, StrategyQA, and ARC-Challenge demonstrate that \FRAMEWORK{} reduces token consumption by 31--53\% while maintaining accuracy, confirming that \METRIC{} successfully distills informational ``froth'' from reasoning chains without compromising deductive validity.
\end{abstract}

\section{Introduction}

Large Language Models (LLMs) have demonstrated remarkable capabilities in complex reasoning tasks, largely driven by the Chain-of-Thought (CoT) prompting strategy~\cite{wei2022chain}. By decomposing complex problems into intermediate steps, CoT helps bridge the gap between question and answer. However, these gains often come at the cost of inference efficiency. Recent studies show that LLMs tend to \textbf{``over-reason''} by generating verbose explanations, repetitive statements, or contextually irrelevant details that increase computational cost without adding deductive value~\cite{turpin2023language, wang2023selfconsistency, chiang-lee-2024-reasoning}. More recent work has therefore begun to treat \emph{reasoning efficiency} itself as an important objective, for example through rationale reduction and concise intermediate reasoning formats~\cite{jang-etal-2025-verbosity, xu2025chain}.

\begin{figure}
    \centering
    \includegraphics[width=\linewidth]{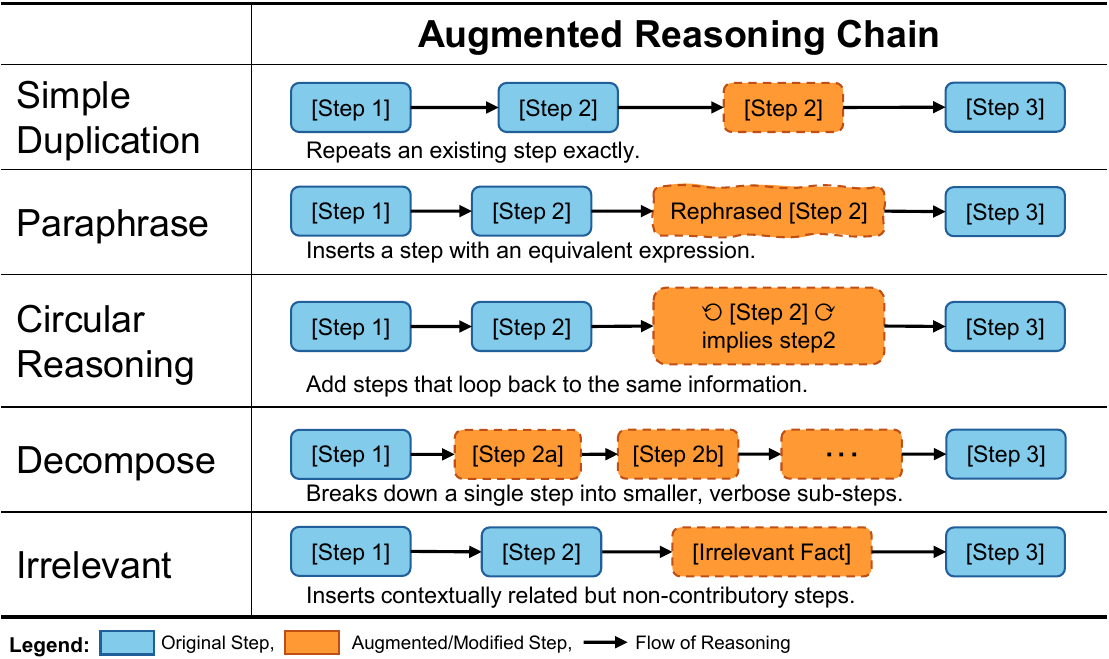}
    \caption{Taxonomy of reasoning inefficiencies in \BENCHMARK. The diagram illustrates how five distinct types of redundant steps are synthetically injected into the reasoning chain to simulate valid but dispensable ``froth.''}
    \label{fig:RIV}
\vspace{-4pt}
\end{figure}

To assess and improve reasoning quality, various Process Reward Models (PRMs) and reasoning-step evaluators have been proposed, including ReasonEval and Math-Shepherd~\cite{Xia_Li_Liu_Wu_Liu_2025, wang-etal-2024-math}. More broadly, recent benchmarks and surveys suggest that reasoning quality extends beyond step correctness to dimensions such as coherence, utility, and simplicity~\cite{song-etal-2025-prmbench, lee-hockenmaier-2025-evaluating}. In parallel, faithfulness studies have shown that final-answer accuracy can sometimes remain high even when parts of a reasoning trace are truncated or perturbed, highlighting the need for careful controls when evaluating compressed CoT traces~\cite{lanham2023measuring}. Despite these developments, existing evaluators remain largely optimized for \textbf{correctness} and \textbf{logical validity}. Our analysis reveals a critical blind spot: they often fail to distinguish \textit{inefficiency} from \textit{reasoning}. In particular, they may assign high scores to ``valid but redundant'' steps---such as excessive decomposition or circular logic---simply because those steps remain factually true and linguistically coherent. As a result, current reasoning evaluation still focuses primarily on ``Is this step true?'', while largely overlooking the equally important question: \textbf{``Is this step truly necessary?''}

In this paper, we shift the focus of reasoning evaluation from correctness alone toward \textbf{information density}. To systematically diagnose the limitations of current evaluators, we first introduce \textbf{\BENCHMARK}, a diagnostic benchmark derived from GSM8K~\cite{cobbe2021trainingverifierssolvemath}. As illustrated in Figure~\ref{fig:RIV}, \BENCHMARK{} injects five distinct types of inefficiency, ranging from simple duplication to subtle circular reasoning. Using this benchmark, we show that existing validity-focused PRMs are largely insensitive to explicit redundancy.

To address this gap, we propose \textbf{\METRIC} (Context-Aware Information Density), a novel unsupervised metric that evaluates reasoning steps using information-utility signals such as local novelty, global goal alignment, and information density. Unlike prior metrics, \METRIC{} is designed to identify informational ``froth'' within reasoning chains without relying on a reference trace. To \textbf{empirically validate the diagnostic precision} of this metric, we introduce \textbf{\FRAMEWORK} (Pruning And Compression for Efficiency), a post-hoc compression strategy. Rather than simply deleting steps, \FRAMEWORK{} identifies \textit{latent inefficiency} in reasoning chains and predominantly \textbf{compresses} verbose steps (Merge) while pruning irrelevant ones, making the chain substantially more compact while preserving logical coherence. To strengthen this validation, we further compare \FRAMEWORK{} against random step-removal controls and PRM-based compression baselines, showing that its gains are not explained by trivial pruning and that validity-oriented evaluators yield only limited compression in the same setting.

Our main contributions are summarized as follows:
\begin{itemize}
    \item We uncover the ``efficiency blind spot'' of current reasoning evaluators through \textbf{\BENCHMARK}, a stress-test benchmark designed to diagnose specific types of reasoning inefficiency.
    \item We propose \textbf{\METRIC}, an interpretable, training-free metric that estimates the informational utility of reasoning steps, distinguishing essential logic from redundant ``froth.''
    \item We validate our approach via \textbf{\FRAMEWORK}, which reduces token consumption by 31--53\% across arithmetic, commonsense, and scientific reasoning tasks with minimal accuracy loss. Additional control experiments with random pruning and PRM-based compression baselines show that these gains arise from \textbf{selective compression}, not trivial deletion. Together, these results provide empirical evidence for substantial \textit{latent inefficiency} in standard CoT reasoning, suggesting that high-quality reasoning data can be considerably more compact than commonly assumed.
\end{itemize}

\section{Related Work}

\subsection{Over-reasoning and Inference Efficiency}
Chain-of-Thought (CoT) prompting~\cite{wei2022chain} has substantially improved multi-step reasoning in LLMs, but often at the cost of inference efficiency. Recent studies report that LLMs frequently produce overly long rationales, repetitive verification loops, or contextually unhelpful details---a phenomenon often described as \textbf{over-reasoning}~\cite{turpin2023language, chen2024frugalgpt, chiang-lee-2024-reasoning}. Such redundancy can increase computational cost and, in some cases, even harm performance by distracting the model from task-relevant information~\cite{jiang2023llmlingua}. More recent work has therefore started to treat \emph{reasoning efficiency} itself as an optimization target, for example by shortening verbose rationales or encouraging concise intermediate reasoning formats~\cite{jang-etal-2025-verbosity, xu2025chain}.

A related line of work aims to reduce inference cost through token pruning, such as H2O~\cite{zhang2023ho} and Learned Token Pruning~\cite{10.1145/3534678.3539260}. These methods operate at the \textbf{token level}, primarily targeting runtime efficiency (e.g., attention or KV-cache reduction). By contrast, our focus is on \textbf{semantic inefficiency within reasoning steps}. Rather than pruning tokens during generation, \FRAMEWORK{} identifies and compresses low-utility reasoning content after generation. This makes our approach complementary to runtime-oriented token pruning methods.

\subsection{Reasoning Step Evaluation Methods}
Beyond outcome-based evaluation, step-wise evaluation methods have been proposed to provide finer-grained supervision for reasoning. Process Reward Models (PRMs), such as PRM800K~\cite{lightman2024lets} and Math-Shepherd~\cite{wang-etal-2024-math}, are primarily designed to distinguish correct from incorrect intermediate reasoning. More recently, \textbf{ReasonEval}~\cite{Xia_Li_Liu_Wu_Liu_2025} extended this direction by considering additional dimensions such as redundancy and clarity. In parallel, broader benchmarks and surveys have argued that reasoning quality should be assessed along multiple axes---including correctness, coherence, utility, and simplicity---rather than correctness alone~\cite{song-etal-2025-prmbench, lee-hockenmaier-2025-evaluating}.

Despite this progress, existing evaluators remain limited for diagnosing inefficiency. Standard PRMs are optimized mainly for \textbf{correctness verification}: they penalize factual or logical errors, but often assign favorable scores to steps that are valid yet unnecessary. ReasonEval moves closer to our goal, but its judgments are learned from supervised annotations, where \emph{necessity} and \emph{conciseness} can be harder to define consistently than correctness. In contrast, our approach uses \textbf{\BENCHMARK{}} as a controlled diagnostic benchmark, where inefficiencies are synthetically injected under explicit constraints. This setup enables a more direct test of whether an evaluator can distinguish \emph{necessary} reasoning from \emph{valid but inefficient} reasoning.

\begin{figure*}[t]
    \centering
    \includegraphics[width=\linewidth]{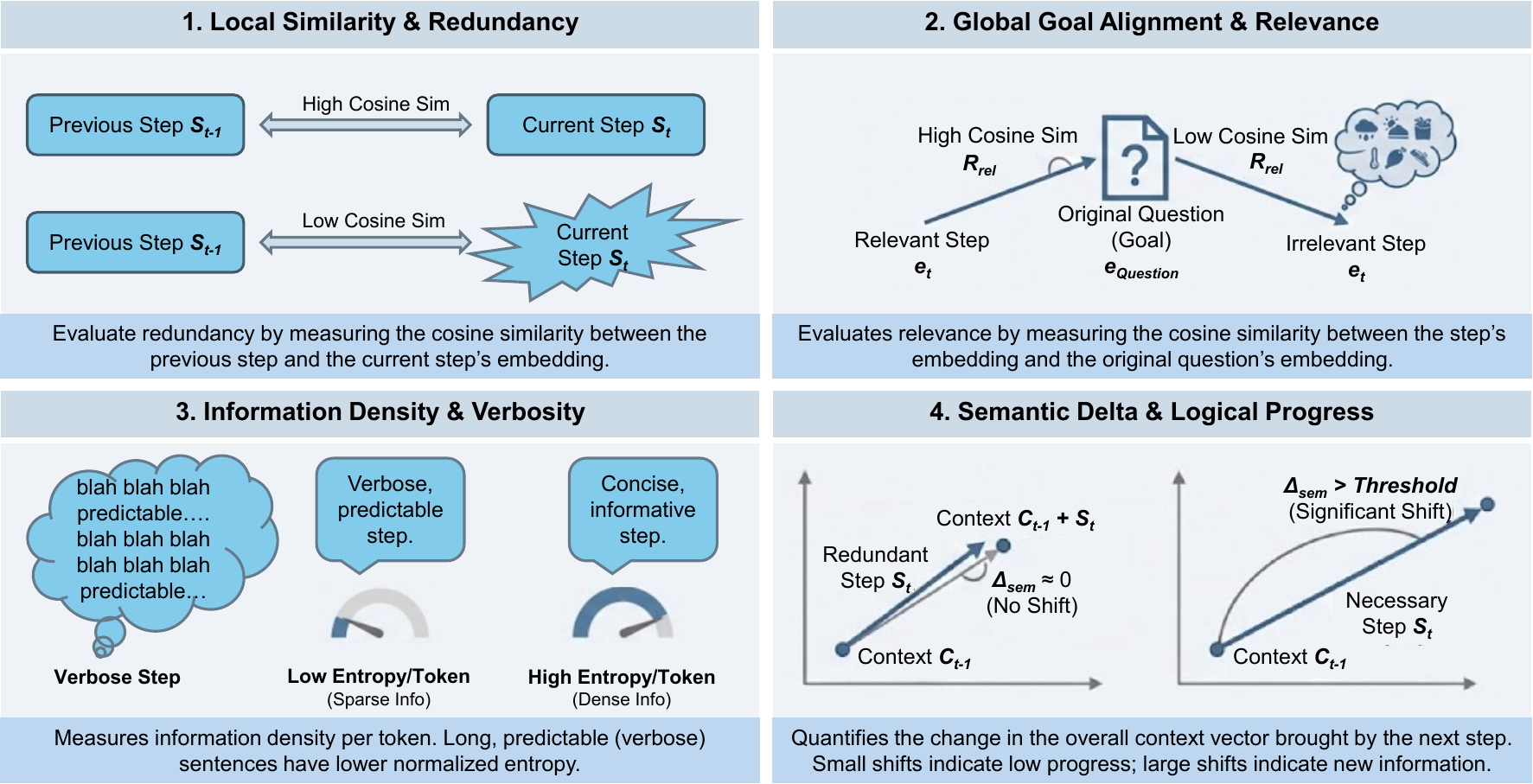}
    \caption{Conceptual overview of the four signals used by CAID. (1) Local Similarity detects surface-level redundancy via adjacent step comparison. (2) Global Goal Alignment filters irrelevant steps by measuring drift from the original question. (3) Information Density identifies verbose, low-entropy steps using length-normalized perplexity. (4) Semantic Delta quantifies the vector shift in the context, ensuring substantial logical progress.}
    \label{fig:CAID_metrics}
\vspace{-6pt}
\end{figure*}

\subsection{Redundancy Detection Metrics}
Several prior works have addressed redundancy or information compression in text. ROSCOE~\cite{golovneva2023roscoe} proposes a suite of metrics for evaluating reasoning quality, including semantic-similarity-based measures for detecting repetition. Similarly, \citet{li2023compressing} introduced Selective Context, which uses self-information (perplexity) to compress prompts by removing less informative content. Related faithfulness studies further showed that final-answer accuracy can sometimes remain high even when portions of a reasoning trace are truncated, highlighting an important caveat for compression-based evaluation of CoT~\cite{lanham2023measuring}.

However, these approaches do not fully address inefficiency in multi-step reasoning. Similarity-based metrics such as ROSCOE are most effective for detecting overt repetition, but are less suited to cases where a step is locally fluent yet contributes little to global progress. Reference-based evaluation is also restrictive in our setting, because multiple reasoning chains may be valid and a reference trace is not necessarily efficiency-optimal. Metrics such as Selective Context capture static information content, but do not explicitly model step-wise relevance or logical progress. \METRIC{} addresses these gaps with a \textbf{reference-free} formulation that combines local redundancy, global goal alignment, information density, and \textbf{Semantic Delta} to estimate the utility of each reasoning step in context.

\section{Methodology}

We present a framework for analyzing and reducing reasoning inefficiency, shifting evaluation from a purely validity-centric perspective toward an \textbf{efficiency-aware} one. Our approach consists of three stages:
(1) \textbf{Diagnosis:} we introduce \textbf{\BENCHMARK{}} to test whether existing evaluators can detect inefficient but valid reasoning steps;
(2) \textbf{Measurement:} we propose \textbf{\METRIC{}}, a reference-free metric for estimating the utility of individual reasoning steps;
and (3) \textbf{Validation:} we apply \textbf{\METRIC{}} within \textbf{\FRAMEWORK{}} to examine whether the identified steps can be compressed without substantially degrading downstream performance.

\subsection{\BENCHMARK: Diagnosing Inefficiency}
\label{sec:riv_construction}

Evaluating reasoning efficiency is challenging because there is rarely a single ground-truth decomposition of which intermediate steps are \emph{necessary}. Human judgments on necessity are often subjective, especially when multiple reasoning traces are valid. To obtain a more controlled testbed, we construct \textbf{\BENCHMARK{}}, a diagnostic benchmark derived from GSM8K~\cite{cobbe2021trainingverifierssolvemath} under a \textbf{relative inefficiency} setting. Starting from baseline CoT traces, we inject perturbations that are \emph{less efficient} than the original steps while remaining factually and logically valid. This yields a controlled setting for testing whether an evaluator can distinguish necessary reasoning from \emph{valid but inefficient} reasoning.

\paragraph{Construction Process.}
The full construction procedure is described in Appendix~\ref{app:construction_details}. We use a hybrid generation strategy to balance control and diversity. \textbf{Simple Duplication} is created by rule-based repetition, while the more complex types---\textbf{Paraphrase}, \textbf{Decompose}, \textbf{Circular}, and \textbf{Irrelevant}---are generated with \textbf{GPT-4o} as the generator $\mathcal{G}$.

To ensure that injected steps remain \emph{valid but inefficient}, we impose two constraints during generation:
\begin{itemize}
    \item \textbf{No New Progress:} the generated step must not advance the reasoning state beyond the target step $S_t$.
    \item \textbf{Contextual Coherence:} especially for the \textit{Irrelevant} type, the generated content must remain mathematically true and locally coherent with the surrounding context, while contributing nothing necessary to the solution.
\end{itemize}

The prompts for each perturbation type are provided in Appendix~\ref{app:prompt_details}. In total, \BENCHMARK{} contains 7,473 samples and over 20,000 injected steps (Appendix~\ref{app:dataset_stats}).

As illustrated in Figure~\ref{fig:RIV}, we define five types of reasoning inefficiency designed to approximate common over-reasoning patterns in LLMs (see Appendix~\ref{app:qualitative_examples} for examples). \textbf{Simple Duplication} and \textbf{Paraphrase} capture lexical and semantic redundancy. \textbf{Decompose} models excessive fragmentation of a single reasoning step into multiple low-utility micro-steps. \textbf{Circular Reasoning} and \textbf{Irrelevant} target stalled logical progress and deviation from the problem objective, respectively.

\paragraph{Human Verification.}
To verify the quality of the synthetic perturbations, we manually evaluated the four GPT-4o-generated types (\textit{Paraphrase}, \textit{Decompose}, \textit{Circular}, and \textit{Irrelevant}), excluding the rule-based \textit{Simple Duplication}. We randomly sampled 30 instances per type (120 total) and checked whether each example satisfied \textit{Logical Equivalence} and the \textit{No New Progress} constraint.

Overall, the perturbations were highly reliable. \textit{Irrelevant} and \textit{Circular Reasoning} achieved near-perfect validity (29/30 and 30/30, respectively). \textit{Decompose} showed a slightly higher failure rate (4/30), mainly for atomic steps (e.g., simple equations such as $3 \times x = 30$) that are difficult to decompose further without introducing hallucinated details or future-step leakage. Despite these edge cases, most generated steps matched the intended inefficiency type without altering the underlying solution logic.

\subsection{\METRIC: Context-Aware Information Density}
\label{sec:caid_metric}

Existing evaluators are largely optimized for factual correctness, and therefore often miss reasoning steps that are valid but inefficient. To address this gap, we propose \textbf{\METRIC{}}, a \textbf{reference-free} and unsupervised metric for estimating the \textit{utility} of a reasoning step in context. Rather than targeting a single error type, \METRIC{} combines four complementary signals that capture redundancy, relevance, information density, and contextual progress. Figure~\ref{fig:CAID_metrics} summarizes these components.

\paragraph{1. Local Similarity (Redundancy).}
To detect local repetition, we measure the cosine similarity between the current step $S_t$ and its immediate predecessor $S_{t-1}$ using a lightweight encoder $\mathcal{E}$:
\[
\mathcal{M}_{sim}(S_t) = \text{CosSim}(\mathcal{E}(S_t), \mathcal{E}(S_{t-1})).
\]
A high similarity score indicates that $S_t$ contributes little beyond the immediately preceding step, which is characteristic of duplication or inefficient paraphrasing.

\paragraph{2. Global Goal Alignment (Relevance).}
Useful reasoning steps should remain aligned with the problem objective. We therefore measure the similarity between the step and the original question $Q$:
\[
\mathcal{M}_{rel}(S_t) = \text{CosSim}(\mathcal{E}(S_t), \mathcal{E}(Q)).
\]
A low alignment score suggests that the step is locally fluent but weakly connected to the goal of the reasoning process.

\paragraph{3. Information Density (Verbosity).}
We measure how much information a step conveys relative to its length using length-normalized perplexity under a causal language model $\mathcal{M}$:
\begin{equation}
    \mathcal{M}_{density}(S_t) =
    \frac{\log(\mathrm{PPL}_{\mathcal{M}}(S_t))}{\mathrm{Length}(S_t)}.
\end{equation}
Low-density steps are typically verbose and predictable relative to their token count. This signal is particularly useful for identifying overly decomposed micro-steps that preserve validity while contributing little new content.

\paragraph{4. Semantic Delta (Logical Progress).}
A reasoning step can be valid yet still fail to move the reasoning state forward. To capture this, we define \textit{Semantic Delta} as the change in the contextual representation after adding $S_t$:
\begin{equation}
    \mathcal{M}_{delta}(S_t) =
    1 - \text{CosSim}(\mathcal{E}(C_{t-1}), \mathcal{E}(C_{t-1} \oplus S_t)),
\end{equation}
where $C_{t-1}$ denotes the accumulated context up to step $t-1$. A near-zero delta indicates that the new step makes little contextual progress, as in tautological restatements or circular verification.

Because later reasoning steps often yield smaller marginal gains as the chain approaches its conclusion, we use a decaying threshold for $\mathcal{M}_{delta}$:
\[
\tau_{\delta}(t) = \tau_{\delta}^{base}\lambda^t,
\]
which reduces sensitivity to small contextual updates in later positions.

\paragraph{Decision Logic.}
\METRIC{} maps these signals to an action set $\mathcal{A}=\{\textsc{Prune}, \textsc{Merge}, \textsc{Keep}\}$. Steps with high local redundancy or low goal alignment are treated as unnecessary and assigned \textsc{Prune}. Steps that remain relevant but have low information density or low semantic delta are assigned \textsc{Merge}, indicating that their content may be useful but should be expressed more compactly. All remaining steps are assigned \textsc{Keep}. This design allows \METRIC{} to distinguish \emph{removable} steps from \emph{compressible} ones, rather than treating all low-utility content as equally disposable.

\subsection{Application: Validating \METRIC{} via \FRAMEWORK{}}
\label{sec:pace_application}

To empirically validate the diagnostic precision of \METRIC{}, we introduce \textbf{\FRAMEWORK} (Pruning And Compression for Efficiency) as a \textbf{post-hoc compression strategy}. Our primary objective here is not to accelerate real-time inference, but to use compression as a diagnostic probe: if the steps flagged by \METRIC{} can be removed or rewritten into denser forms without harming downstream reasoning, this provides evidence that they were inefficient in their original form.

\FRAMEWORK{} operates in a \textit{Generate-then-Refine} pipeline. Based on the classification from \METRIC{}, we apply three actions:
\begin{itemize}
    \item \textbf{\textsc{Prune}:} Removes steps flagged as highly redundant ($\mathcal{M}_{sim}$) or weakly relevant to the problem goal ($\mathcal{M}_{rel}$).
    
    \item \textbf{\textsc{Merge}:} Compresses steps exhibiting \textit{latent inefficiency} (i.e., valid but overly verbose or fragmented) using an LLM re-writer. To prevent semantic drift or information overload, we enforce two safety constraints before merging step $S_t$ into the accumulated step $S'_{last}$:
    \begin{enumerate}
        \item \textbf{\mbox{Semantic Consistency:}}
        $\text{CosSim}(\mathcal{E}(S'_{last}), \mathcal{E}(S_t)) \ge \tau_{merge}$.
        This ensures that the new content remains logically compatible with the current context.
        \item \textbf{\mbox{Information Saturation:}}
        $\mathcal{I}(S'_{last}) \le \tau_{max}$.
        This prevents merging when the current accumulated step is already sufficiently information-dense, thereby avoiding readability loss.
    \end{enumerate}
    If either constraint is violated, the merge is halted and a new step is initiated.
    
    \item \textbf{\textsc{Keep}:} Retains steps that appear necessary for logical progress.
\end{itemize}

\paragraph{Addressing the ``Trivial Accuracy'' Concern.}
One might assume that maintaining accuracy is trivial if the final conclusion step is preserved. To reduce this possibility, we construct the evaluation prompt using the compressed chain $C'$ while \textbf{excluding the final answer} (i.e., the ``\#\#\# Result'' token). The model must therefore \textit{regenerate} the final answer solely from the remaining reasoning trace. Since reasoning chains are causal, removing or altering a genuinely necessary intermediate step should break the dependencies required to derive the correct solution. In this sense, answer preservation after compression serves as evidence that the removed steps were not essential in their original form. In the experiments, we further strengthen this evaluation by comparing \FRAMEWORK{} against non-selective pruning controls, allowing us to distinguish selective compression from trivial answer recoverability.

\begin{table*}[t]
\centering
\resizebox{\textwidth}{!}{
\begin{tabular}{l|cc|cc|cc|cc|cc}
\toprule
\multirow{2}{*}{Model} &
\multicolumn{2}{c|}{Simple Duplication} &
\multicolumn{2}{c|}{Paraphrase} &
\multicolumn{2}{c|}{Decompose} &
\multicolumn{2}{c|}{Circular Reasoning} &
\multicolumn{2}{c}{Irrelevant} \\
& Aug PR ($\downarrow$) & Gold PR & Aug PR ($\downarrow$) & Gold PR & Aug PR ($\downarrow$) & Gold PR & Aug PR ($\downarrow$) & Gold PR & Aug PR ($\downarrow$) & Gold PR \\
\midrule
ReasonEval 7B & 0.6555 & \textbf{0.9897} & 0.7338 & \textbf{0.9867} & 0.7917 & \textbf{0.9853} & 0.4809 & \textbf{0.9746} & \underline{0.1509} & \textbf{0.9571} \\
ReasonEval 34B & 0.7779 & 0.9558 & 0.7251 & \underline{0.9533} & 0.7604 & \underline{0.9571} & \underline{0.3762} & 0.9345 & \textbf{0.0331} & 0.9118 \\
\cmidrule(lr){1-11}
ThinkPRM 1.5B & \underline{0.6264} & 0.7342 & \underline{0.6821} & 0.7486 & \underline{0.7187} & 0.7663 & 0.7179 & 0.7810 & 0.6667 & 0.7353 \\
ThinkPRM 7B & 0.6849 & 0.8003 & 0.7834 & 0.8149 & 0.7326 & 0.8046 & 0.7619 & 0.8881 & 0.7999 & 0.8682 \\
ThinkPRM 14B & 0.8582 & 0.9142 & 0.8859 & 0.9140 & 0.8150 & 0.8706 & 0.9118 & 0.9274 & 0.8474 & 0.9188 \\
\cmidrule(lr){1-11}
Qwen2.5-Math-PRM-7B & 0.9679 & \underline{0.9606} & 0.9512 & 0.9521 & 0.9350 & 0.9433 & 0.8614 & \underline{0.9562} & 0.9746 & \underline{0.9512} \\
Qwen2.5-Math-PRM-72B & 0.8368 & 0.9517 & 0.8896 & 0.9457 & 0.9108 & 0.9502 & 0.8703 & 0.9554 & 0.8961 & 0.9504 \\
\hline \hline
CAID (Ours) & \textbf{0.0000} & 0.5752 & \textbf{0.0174} & 0.5596 & \textbf{0.2006} & 0.5142 & \textbf{0.0190} & 0.4799 & 0.1598 & 0.5155 \\
\bottomrule
\end{tabular}
}
\caption{Step Preservation Rate (SPR) comparison by augmentation type. Aug PR: Augmented Step Preservation Rate (lower is better), Gold PR: Gold Step Preservation Rate (higher indicates retention). \textbf{Bold} indicates the best performance, and \underline{underlined} indicates the second-best performance.}
\label{tab:aug_type_results}
\end{table*}

\section{Experiments}

\subsection{Experimental Setup}

\paragraph{Datasets.}
We employ diverse benchmarks to conduct a two-stage evaluation, assessing both diagnostic sensitivity and practical compression utility.
\begin{itemize}
    \item \textbf{\BENCHMARK{}:} A controlled diagnostic set used to measure the sensitivity of metrics to explicit, synthetically injected inefficiencies (Section~\ref{sec:riv_construction}).
    \item \textbf{Standard Benchmarks:} To validate \FRAMEWORK{} on real-world reasoning, we use \textbf{GSM8K}~\cite{cobbe2021trainingverifierssolvemath}, \textbf{StrategyQA}~\cite{geva2021strategyqa}, and \textbf{ARC-Challenge}~\cite{DBLP:journals/corr/abs-1803-05457}. These datasets cover arithmetic, commonsense, and scientific reasoning, respectively, allowing us to test whether \METRIC{} generalizes across different reasoning domains without task-specific tuning.
\end{itemize}

\paragraph{Baselines.}
For the diagnostic comparison on \BENCHMARK{}, we evaluate \textbf{multiple scales} of representative PRM-based evaluators, including \textbf{ReasonEval}, \textbf{ThinkPRM}, and \textbf{Qwen2.5-Math-PRM}, alongside \METRIC{}. For compression validation via \FRAMEWORK{}, we use \textbf{Llama-3.1-8B-Instruct} as the backbone model and compare compressed chains against the standard \textbf{Zero-shot CoT} baseline to measure the trade-off between token reduction and answer accuracy on GSM8K, StrategyQA, and ARC-Challenge. To test whether the gains of \FRAMEWORK{} arise from \emph{selective compression} rather than trivial answer recoverability, we additionally include \textbf{Remove-Last} and \textbf{Random Pruning} controls on \textbf{GSM8K}, following prior faithfulness concerns~\cite{lanham2023measuring}. We further compare against \textbf{PRM-based compression baselines} on \textbf{GSM8K} by replacing \METRIC{} with \textbf{ThinkPRM-14B} and \textbf{Qwen2.5-Math-PRM-7B} within the same \FRAMEWORK{} pipeline, thereby isolating the effect of the evaluator while keeping the rewriting and answer-regeneration procedure fixed.

\paragraph{Implementation of \METRIC{}.}
We implement \METRIC{} using lightweight off-the-shelf models: \textbf{all-MiniLM-L6-v2} (22M) for semantic encoding and \textbf{GPT-2 Small} (124M) for density estimation. We use a fixed set of hyperparameters across all datasets without task-specific tuning to test robustness and cross-domain transferability. Detailed model configurations, threshold values, and sensitivity analyses are provided in Appendix~\ref{app:implementation_details}.

\subsection{Results 1: Diagnostic Capability on \BENCHMARK{}}

We evaluate how well different evaluators handle the inefficiencies injected into \BENCHMARK{}. Instead of binary accuracy, we report the \textbf{Step Preservation Rate (SPR)}, defined as the ratio of steps retained after evaluation.
\begin{itemize}
    \item \textbf{Augmented PR (Aug PR):} Measures recall on inefficient steps. \textbf{Lower is better}, indicating the model successfully removed or flagged the inefficient step.
    \item \textbf{Gold PR:} Measures retention of original human-written steps. \textbf{Higher typically indicates safety}, assuming human steps are perfectly efficient. However, as discussed below, we challenge this assumption.
\end{itemize}

\paragraph{Blind Spot of Validity-Focused Evaluators.}
As shown in Table~\ref{tab:aug_type_results}, existing methods exhibit surprisingly high Aug PR scores. A critical finding is the performance of \textbf{ReasonEval}. Despite being a specialized reasoning step evaluator equipped with an explicit \textit{redundancy score}, it fails to effectively penalize redundant steps, retaining approximately 70\% of \textit{Simple Duplication}, \textit{Paraphrase}, and \textit{Decompose} types. Similarly, even the 72B-parameter Qwen2.5-Math-PRM preserves 83.68\% of \textit{Simple Duplications}. This confirms that validity-focused models, regardless of their size or specific scoring sub-metrics, remain essentially blind to inefficiency as long as the statement is factually correct.

\paragraph{Effectiveness and Efficiency of \METRIC{}.}
In contrast, \METRIC{} achieves near-perfect detection on redundancy, with an Aug PR of \textbf{0.0000} for Duplication and \textbf{0.0174} for Paraphrasing. It also effectively identifies complex inefficiencies like Circular Reasoning (0.0190) and Decomposition (0.2006), where baselines struggle significantly.
Regarding \textit{Irrelevant} steps, while the large-scale supervised model \textbf{ReasonEval-34B} achieves the best performance (0.0331), \METRIC{} (0.1598) demonstrates competitive capability, performing comparably to \textbf{ReasonEval-7B} (0.1509). Crucially, \METRIC{} achieves this with a total of only \textbf{146M parameters} (124M GPT-2 + 22M MiniLM), whereas ReasonEval requires 7B to 34B parameters. This demonstrates that \METRIC{} delivers robust diagnostic precision with orders of magnitude greater computational efficiency than large-scale supervised evaluators.

\paragraph{Redefining ``Gold'': Deletion vs. Compression.}
A distinct characteristic of \METRIC{} is its lower Gold PR ($\approx$0.55) compared to baselines ($>0.90$). While this might initially appear as over-penalization, a granular analysis of the action distribution reveals that \METRIC{} is not ``wrong,'' but rather stricter regarding information density.
Out of 11,899 Gold steps not fully preserved by \METRIC{}:
\begin{itemize}
    \item Only \textbf{1.5\% (184 steps)} were flagged for removal (\textsc{Prune}), primarily due to high redundancy (169 steps) or irrelevance (15 steps).
    \item The remaining \textbf{98.5\% ($\approx$11,700 steps)} were flagged for \textbf{\textsc{Merge}}.
\end{itemize}
As qualitatively analyzed in Appendix~\ref{app:qualitative_analysis_gold} (Table~\ref{tab:gold_merge_examples}), these flagged steps are factually valid but functionally inefficient. For instance, steps that merely restate a calculated value (e.g., ``Child = 4'' $\rightarrow$ ``Ticket is \$4'') trigger the \textbf{Low Semantic Delta} criteria due to a lack of logical progress. Similarly, steps that verbally describe an operation before executing it (e.g., ``Then multiply the number...'') are flagged for \textbf{Low Information Density}. This empirical evidence suggests that standard datasets contain significant \textbf{latent inefficiency}, validating our approach of \textit{compression} over blind retention.

\subsection{Results 2: Efficiency via \FRAMEWORK{}}

We next evaluate whether the steps identified by \METRIC{} can be compressed while preserving downstream performance. To this end, we apply \FRAMEWORK{} to reasoning chains generated by Llama-3.1-8B. Table~\ref{tab:pace_results} summarizes the trade-off between answer accuracy and token reduction on standard reasoning benchmarks. To further test whether these gains reflect \emph{selective compression} rather than trivial answer recoverability, we additionally compare \FRAMEWORK{} against non-selective pruning controls and PRM-based compression baselines on GSM8K.

\begin{table}[h]
\centering
\resizebox{\linewidth}{!}{
\begin{tabular}{l|c|cc|cc}
\hline
\multirow{2}{*}{\textbf{Dataset}} & \multirow{2}{*}{\textbf{Method}} & \multicolumn{2}{c|}{\textbf{Performance}} & \multicolumn{2}{c}{\textbf{Efficiency}} \\
 & & \textbf{Acc (\%)} & \textbf{$\Delta$} & \textbf{Tokens} & \textbf{Red (\%)} \\ \hline
\multirow{2}{*}{GSM8K} & Baseline & 82.03 & - & 214.6 & - \\
 & \textbf{\FRAMEWORK} & 81.12 & -0.91 & \textbf{148.0} & \textbf{-31.0\%} \\ \hline
\multirow{2}{*}{StrategyQA} & Baseline & 70.31 & - & 327.7 & - \\
 & \textbf{\FRAMEWORK} & 69.93 & -0.37 & \textbf{154.4} & \textbf{-52.9\%} \\ \hline
\multirow{2}{*}{ARC-C} & Baseline & 83.70 & - & 277.8 & - \\
 & \textbf{\FRAMEWORK} & \textbf{84.64} & \textbf{+0.94} & \textbf{156.7} & \textbf{-43.6\%} \\ \hline
\end{tabular}
}
\caption{Comparison of accuracy and token usage between baseline CoT and \FRAMEWORK{}. \FRAMEWORK{} significantly reduces tokens while maintaining or improving accuracy.}
\label{tab:pace_results}
\end{table}

\paragraph{Compression with Limited Accuracy Loss.}
Across GSM8K, StrategyQA, and ARC-Challenge, \FRAMEWORK{} reduces token usage by 31.0\%--52.9\% while maintaining similar answer accuracy. These results suggest that standard CoT traces contain a substantial amount of \textbf{compressible} content: many generated steps appear to be useful enough to preserve in condensed form, but not necessary in their original verbose form.

\paragraph{Selective Compression, Not Trivial Pruning.}
A natural concern is that final-answer accuracy may remain high after pruning simply because the model can reconstruct the answer from a partially preserved trace. To test this, we compare \FRAMEWORK{} against \textit{Remove-Last} and \textit{Random Pruning} controls. As shown in Table~\ref{tab:pace_random_controls}, selective compression is substantially more robust than non-selective deletion: \FRAMEWORK{} retains 81.12\% accuracy with 31.05\% token reduction, whereas random pruning at a comparable compression level leads to much larger degradation (e.g., 66.72\% accuracy under 50\% random pruning). Even removing only the final step lowers accuracy to 80.06\%, suggesting that performance is not explained by simple answer copying from the last reasoning step alone. Overall, these results indicate that \FRAMEWORK{} preserves key dependencies in the reasoning chain while compressing low-utility content.

\begin{table}[t]
\centering
\resizebox{\linewidth}{!}{
\begin{tabular}{l|c|c|c|c}
\hline
\textbf{Method} & \textbf{Step Red. (\%)} & \textbf{Tok Red. (\%)} & \textbf{Acc (\%)} & \textbf{$\Delta$ Acc} \\
\hline
Remove Last Only & -- & 0.26 & 80.06 & -1.97 \\
Random Pruning (30\%) & 30.00 & 21.40 & 76.88 & -5.16 \\
Random Pruning (50\%) & 50.00 & 32.86 & 66.72 & -15.31 \\
Remove Last + Random (50\%) & 50.00 & 29.42 & 62.55 & -19.48 \\
\textbf{\FRAMEWORK{} (Ours)} & 53.53 & 31.05 & 81.12 & -0.91 \\
\hline
\end{tabular}
}
\caption{Comparison between \FRAMEWORK{} and non-selective pruning controls on GSM8K. At comparable compression rates, random pruning causes substantial accuracy degradation.}
\label{tab:pace_random_controls}
\end{table}

\paragraph{Comparison with PRM-based Compression Baselines.}
We further replace \METRIC{} with strong PRM evaluators within the same \FRAMEWORK{} pipeline, using ThinkPRM-14B and Qwen2.5-Math-PRM-7B as scoring modules. This comparison isolates the role of the evaluator while keeping the rewriting and answer-regeneration procedure fixed. As shown in Table~\ref{tab:pace_prm_baselines}, PRM-based variants can maintain competitive answer accuracy, but achieve only limited token reduction ($<8\%$), whereas \METRIC{} yields 31.05\% compression on GSM8K. This result is consistent with our main claim: evaluators optimized primarily for validity are less effective at identifying the redundancy/utility dimension targeted by \METRIC{}.

\begin{table}[t]
\centering
\resizebox{\linewidth}{!}{
\begin{tabular}{l|c|c|c}
\hline
\textbf{Evaluator in \FRAMEWORK{}} & \textbf{Acc (\%)} & \textbf{$\Delta$ Acc} & \textbf{Tok Red. (\%)} \\
\hline
Baseline (Zero-shot) & 82.03 & -- & -- \\
ThinkPRM-14B & 79.83 & -2.20 & 7.20 \\
Qwen2.5-Math-PRM-7B & 83.02 & +0.99 & 5.87 \\
\textbf{\METRIC{} (Ours)} & 81.12 & -0.91 & \textbf{31.05} \\
\hline
\end{tabular}
}
\caption{PRM-based compression baselines on GSM8K within the same \FRAMEWORK{} pipeline. Strong PRM evaluators can maintain reasonably strong answer accuracy, but yield only marginal compression.}
\label{tab:pace_prm_baselines}
\end{table}

\paragraph{A Case of Accuracy Improvement.}
On ARC-Challenge, \FRAMEWORK{} improves accuracy by +0.94 points while reducing tokens by 43.6\%. One plausible explanation is that, in information-heavy scientific reasoning, removing tangential or overly verbose steps can make the remaining context easier to use. We view this as suggestive evidence that compression can sometimes improve clarity, rather than merely reducing cost.

\subsection{Ablation Study}

\begin{figure}[h]
    \centering
    \includegraphics[width=\linewidth]{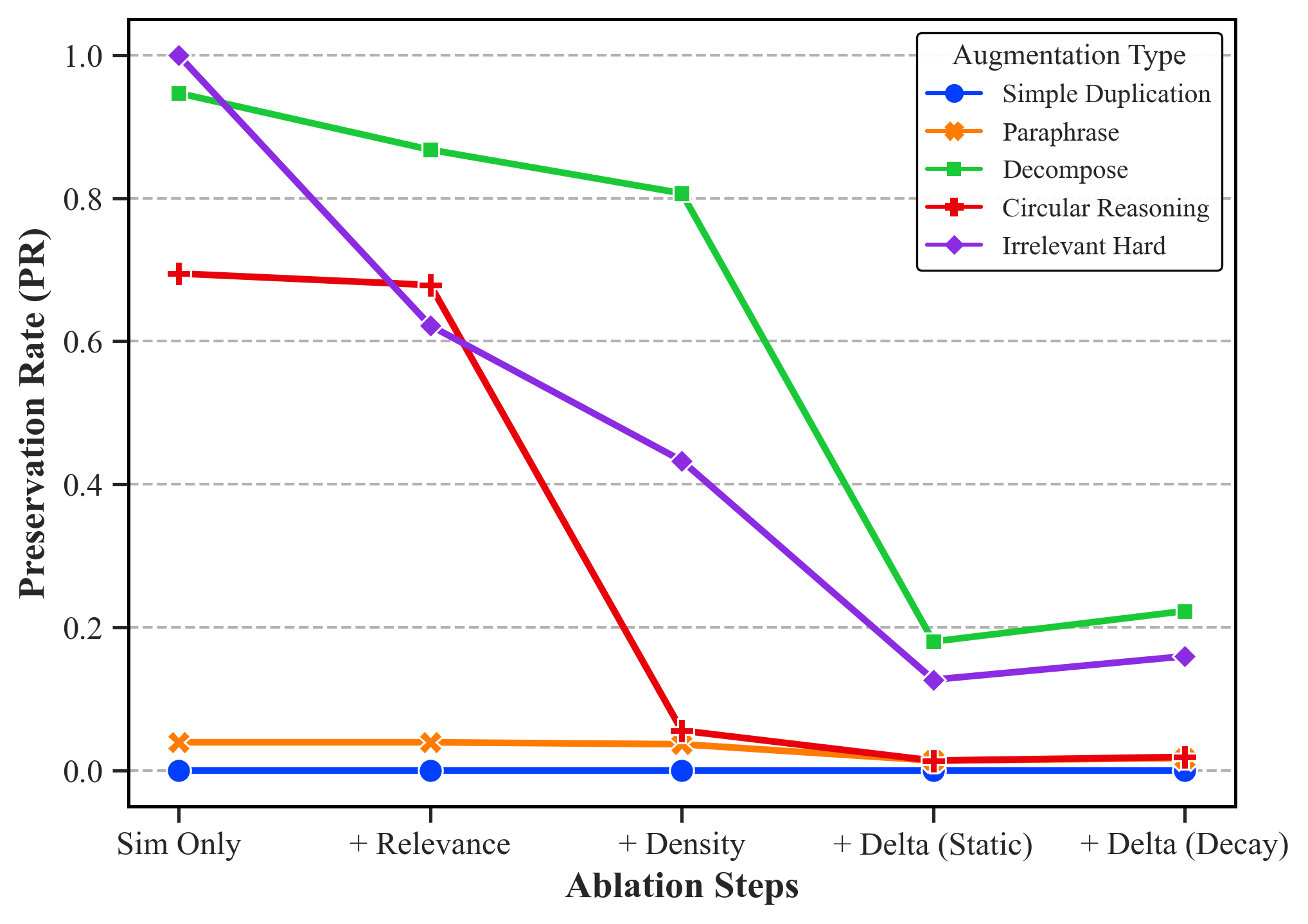}
    \caption{Cumulative effect of CAID components on preservation rate. Lower PR indicates better detection.}
    \label{fig:CAID_ablation_chart}
\vspace{-4pt}
\end{figure}

We next analyze the contribution of each component of \METRIC{} on GSM8K. Figure~\ref{fig:CAID_ablation_chart} shows the cumulative change in preservation rate as each signal is added.

\paragraph{Importance of Semantic Delta.}
As shown in Figure~\ref{fig:CAID_ablation_chart}, \textit{Local Similarity} alone captures overt repetition but is insufficient for more subtle inefficiencies such as \textit{Decomposition} and \textit{Irrelevance}. Adding \textit{Semantic Delta} substantially improves detection, indicating that contextual progress is important for distinguishing genuinely useful steps from those that merely restate or locally elaborate prior content.

\paragraph{Role of Information Density.}
\textit{Information Density} is particularly helpful for \textit{Circular Reasoning}. Tautological or overly predictable steps tend to have low normalized surprisal relative to their length, allowing \METRIC{} to identify verbose but low-utility reasoning that is not captured by similarity alone.

\paragraph{Why Compression Requires More than Deletion.}
We also evaluate different compression actions and safety constraints. Simply removing steps with low logical progress causes a substantial accuracy drop (-5.38\%), suggesting that some low-progress steps still serve as connective structure in the chain. In contrast, \FRAMEWORK{} recovers much of this performance by merging such steps under safety constraints (Consistency and Saturation), supporting the distinction between \textsc{Prune} and \textsc{Merge}. Detailed results are provided in Appendix~\ref{app:pace_ablation}.

\section{Discussion}

\paragraph{Compressible Structure in CoT Traces.}
One notable result is that \METRIC{} marks roughly half of the human-written reference steps as \textbf{compressible} rather than strictly removable. As discussed in Section~4.2, most of these cases are assigned to \textsc{Merge} rather than \textsc{Prune}, indicating that the issue is usually not factual incorrectness or irrelevance, but \textbf{how the reasoning is expressed}. In many cases, valid steps appear verbose, weakly progressive, or fragmented into micro-steps that could be stated more compactly. This suggests that human-written CoT traces are not always efficiency-optimal, even when they are logically sound.

\paragraph{Compression Requires a Different Signal from Validity.}
Our control experiments further clarify this distinction. Random pruning substantially degrades accuracy at similar compression levels, while replacing \METRIC{} with strong PRM evaluators inside the same \FRAMEWORK{} pipeline yields only limited compression. Taken together, these results suggest that \textbf{correctness-oriented evaluation} and \textbf{efficiency-oriented compression} are related but distinct objectives. Existing PRMs are useful for filtering incorrect reasoning, but appear less sensitive to verbose yet valid steps that can be compressed without materially harming downstream performance.

\paragraph{Implications and Limitations.}
Although \FRAMEWORK{} is introduced here as a post-hoc compression method, the results suggest two broader directions. First, compressed traces may be useful for constructing denser reasoning datasets for training or distillation. Second, offline compression could help reduce reasoning-context cost in retrieval-based systems. At the same time, our evaluation remains answer-level: it does not test whether compressed traces preserve the model's full continuation trajectory. A fuller account of redundancy in CoT should therefore examine not only final-answer robustness, but also how intermediate continuations change after compression.

\section{Conclusion}

We studied reasoning inefficiency in Chain-of-Thought (CoT) generation, focusing on cases where intermediate steps are valid yet unnecessarily verbose, repetitive, or weakly connected to the problem objective. Our results suggest that existing reasoning-step evaluators, while effective at verifying correctness, are less effective at identifying this type of low-utility reasoning content.

To address this gap, we introduced three components: \textbf{\BENCHMARK{}}, a controlled benchmark for diagnosing inefficient but valid reasoning steps; \textbf{\METRIC{}}, a reference-free metric for estimating the utility of reasoning steps in context; and \textbf{\FRAMEWORK{}}, a post-hoc compression procedure for testing whether the identified steps can be removed or compressed without substantially harming downstream performance.

Across arithmetic, commonsense, and scientific reasoning benchmarks, \FRAMEWORK{} reduces token usage by 31--53\% while maintaining similar answer accuracy. Additional control experiments show that these gains are not explained by trivial pruning, and that strong PRM-based evaluators achieve only limited compression within the same pipeline. Taken together, these findings suggest that standard CoT traces often contain a meaningful amount of compressible content. More broadly, our results motivate reasoning evaluation that considers not only correctness, but also efficiency and utility.

\section*{Limitations}

Our study has several limitations.

\paragraph{Post-hoc Overhead.}
\FRAMEWORK{} follows a generate-then-refine procedure and therefore does not reduce the latency of the initial inference pass. As a result, it is better suited to offline settings, such as dataset construction or context compression, than to real-time acceleration.

\paragraph{Dependence on Rewriter Quality.}
The \textsc{Merge} action depends on the ability of the underlying LLM to rewrite verbose steps without losing important information. Although our safety constraints ($\tau_{merge}$ and $\tau_{sat}$) help reduce semantic drift, smaller models may still oversimplify complex reasoning during compression.

\paragraph{Domain Scope.}
We evaluate arithmetic (GSM8K), commonsense (StrategyQA), and scientific reasoning (ARC-Challenge). The notion of efficiency may differ in more open-ended domains, such as creative writing, where verbosity can serve stylistic or communicative purposes.

\paragraph{Hyperparameter Transfer.}
Although \METRIC{} is reasonably stable across our benchmarks with a fixed set of thresholds, transferring it to domains with substantially different linguistic characteristics (e.g., code or legal text) may require additional calibration.

\section*{Ethics Statement}

This work is motivated in part by the goal of improving the efficiency of LLM reasoning, which may help reduce unnecessary computational cost. At the same time, we note several ethical considerations.

\paragraph{Data and Privacy.}
All experiments use publicly available datasets. Our study does not involve private or personally identifiable information. The synthetic perturbations in \BENCHMARK{} were generated with GPT-4o for research purposes and were designed to remain task-relevant and non-harmful.

\paragraph{Potential Biases.}
Efficiency-oriented metrics may penalize linguistic styles that are longer or more elaborative by convention. As a result, care is needed when applying \METRIC{} beyond the benchmark settings considered here, especially in contexts where verbosity may reflect dialectal, stylistic, or communicative variation rather than inefficiency.

\paragraph{Use of AI Assistants.}
GPT-4o was used in the data augmentation pipeline to generate synthetic reasoning perturbations for \BENCHMARK{}. AI assistants were also used for preliminary code drafting and language editing. The authors reviewed all such outputs and take full responsibility for the final manuscript.

\section*{Acknowledgments}

This work was supported by Institute of Information \& communications Technology Planning \& Evaluation (IITP) grant funded by the Korea government (MSIT) (RS-2022-00143911, AI Excellence Global Innovative Leader Education Program).

\clearpage

\bibliography{custom}

@inproceedings{
wei2022chain,
title={Chain of Thought Prompting Elicits Reasoning in Large Language Models},
author={Jason Wei and Xuezhi Wang and Dale Schuurmans and Maarten Bosma and Brian Ichter and Fei Xia and Ed H. Chi and Quoc V Le and Denny Zhou},
booktitle={Advances in Neural Information Processing Systems},
editor={Alice H. Oh and Alekh Agarwal and Danielle Belgrave and Kyunghyun Cho},
year={2022},
url={https://openreview.net/forum?id=_VjQlMeSB_J}
}

@inproceedings{chiang-lee-2024-reasoning,
    title = "Over-Reasoning and Redundant Calculation of Large Language Models",
    author = "Chiang, Cheng-Han  and
      Lee, Hung-yi",
    editor = "Graham, Yvette  and
      Purver, Matthew",
    booktitle = "Proceedings of the 18th Conference of the European Chapter of the Association for Computational Linguistics (Volume 2: Short Papers)",
    month = mar,
    year = "2024",
    address = "St. Julian{'}s, Malta",
    publisher = "Association for Computational Linguistics",
    url = "https://aclanthology.org/2024.eacl-short.15/",
    doi = "10.18653/v1/2024.eacl-short.15",
    pages = "161--169"
}

@inproceedings{
wang2023selfconsistency,
title={Self-Consistency Improves Chain of Thought Reasoning in Language Models},
author={Xuezhi Wang and Jason Wei and Dale Schuurmans and Quoc V Le and Ed H. Chi and Sharan Narang and Aakanksha Chowdhery and Denny Zhou},
booktitle={The Eleventh International Conference on Learning Representations },
year={2023},
url={https://openreview.net/forum?id=1PL1NIMMrw}
}

@inproceedings{
turpin2023language,
title={Language Models Don't Always Say What They Think: Unfaithful Explanations in Chain-of-Thought Prompting},
author={Miles Turpin and Julian Michael and Ethan Perez and Samuel R. Bowman},
booktitle={Thirty-seventh Conference on Neural Information Processing Systems},
year={2023},
url={https://openreview.net/forum?id=bzs4uPLXvi}
}

@inproceedings{wang-etal-2024-math,
    title = "Math-Shepherd: Verify and Reinforce {LLM}s Step-by-step without Human Annotations",
    author = "Wang, Peiyi  and
      Li, Lei  and
      Shao, Zhihong  and
      Xu, Runxin  and
      Dai, Damai  and
      Li, Yifei  and
      Chen, Deli  and
      Wu, Yu  and
      Sui, Zhifang",
    editor = "Ku, Lun-Wei  and
      Martins, Andre  and
      Srikumar, Vivek",
    booktitle = "Proceedings of the 62nd Annual Meeting of the Association for Computational Linguistics (Volume 1: Long Papers)",
    month = aug,
    year = "2024",
    address = "Bangkok, Thailand",
    publisher = "Association for Computational Linguistics",
    url = "https://aclanthology.org/2024.acl-long.510/",
    doi = "10.18653/v1/2024.acl-long.510",
    pages = "9426--9439"
}

@article{Xia_Li_Liu_Wu_Liu_2025, title={Evaluating Mathematical Reasoning Beyond Accuracy}, volume={39}, url={https://ojs.aaai.org/index.php/AAAI/article/view/34987}, DOI={10.1609/aaai.v39i26.34987}, abstractNote={The leaderboard of Large Language Models (LLMs) in mathematical tasks has been continuously updated. However, the majority of evaluations focus solely on the final results, neglecting the quality of the intermediate steps. This oversight can mask underlying problems, such as logical errors or unnecessary steps in the reasoning process. To measure reasoning beyond final-answer accuracy, we introduce ReasonEval, a new methodology for evaluating the quality of reasoning steps. ReasonEval employs validity and redundancy to characterize the reasoning quality, as well as accompanying LLMs to assess them automatically. We explore different design options for the LLM-based evaluators and empirically demonstrate that ReasonEval, when instantiated with base models possessing strong mathematical knowledge and trained with high-quality labeled data, consistently outperforms baseline methods in the meta-evaluation datasets. We also highlight the strong generalization capabilities of ReasonEval. By utilizing ReasonEval to evaluate LLMs specialized in math, we find that an increase in final-answer accuracy does not necessarily guarantee an improvement in the overall quality of the reasoning steps for challenging mathematical problems. Additionally, we observe that ReasonEval can play a significant role in data selection. We open-source the best-performing model, meta-evaluation script, and all evaluation results to facilitate future research.}, number={26}, journal={Proceedings of the AAAI Conference on Artificial Intelligence}, author={Xia, Shijie and Li, Xuefeng and Liu, Yixin and Wu, Tongshuang and Liu, Pengfei}, year={2025}, month={Apr.}, pages={27723-27730} }

@misc{cobbe2021trainingverifierssolvemath,
      title={Training Verifiers to Solve Math Word Problems}, 
      author={Karl Cobbe and Vineet Kosaraju and Mohammad Bavarian and Mark Chen and Heewoo Jun and Lukasz Kaiser and Matthias Plappert and Jerry Tworek and Jacob Hilton and Reiichiro Nakano and Christopher Hesse and John Schulman},
      year={2021},
      eprint={2110.14168},
      archivePrefix={arXiv},
      primaryClass={cs.LG},
      url={https://arxiv.org/abs/2110.14168}, 
}

@inproceedings{
lightman2024lets,
title={Let's Verify Step by Step},
author={Hunter Lightman and Vineet Kosaraju and Yuri Burda and Harrison Edwards and Bowen Baker and Teddy Lee and Jan Leike and John Schulman and Ilya Sutskever and Karl Cobbe},
booktitle={The Twelfth International Conference on Learning Representations},
year={2024},
url={https://openreview.net/forum?id=v8L0pN6EOi}
}

@inproceedings{
golovneva2023roscoe,
title={{ROSCOE}: A Suite of Metrics for Scoring Step-by-Step Reasoning},
author={Olga Golovneva and Moya Peng Chen and Spencer Poff and Martin Corredor and Luke Zettlemoyer and Maryam Fazel-Zarandi and Asli Celikyilmaz},
booktitle={The Eleventh International Conference on Learning Representations },
year={2023},
url={https://openreview.net/forum?id=xYlJRpzZtsY}
}

@article{geva2021strategyqa,
  title = {{Did Aristotle Use a Laptop? A Question Answering Benchmark with Implicit Reasoning Strategies}},
  author = {Geva, Mor and Khashabi, Daniel and Segal, Elad and Khot, Tushar and Roth, Dan and Berant, Jonathan},
  journal = {Transactions of the Association for Computational Linguistics (TACL)},
  year = {2021},
}

@article{DBLP:journals/corr/abs-1803-05457,
  author       = {Peter Clark and
                  Isaac Cowhey and
                  Oren Etzioni and
                  Tushar Khot and
                  Ashish Sabharwal and
                  Carissa Schoenick and
                  Oyvind Tafjord},
  title        = {Think you have Solved Question Answering? Try ARC, the {AI2} Reasoning
                  Challenge},
  journal      = {CoRR},
  volume       = {abs/1803.05457},
  year         = {2018},
  url          = {http://arxiv.org/abs/1803.05457},
  eprinttype    = {arXiv},
  eprint       = {1803.05457},
  bibsource    = {dblp computer science bibliography, https://dblp.org}
}

@article{
chen2024frugalgpt,
title={Frugal{GPT}: How to Use Large Language Models While Reducing Cost and Improving Performance},
author={Lingjiao Chen and Matei Zaharia and James Zou},
journal={Transactions on Machine Learning Research},
issn={2835-8856},
year={2024},
url={https://openreview.net/forum?id=cSimKw5p6R},
note={Featured Certification}
}

@inproceedings{
jiang2023llmlingua,
title={{LLML}ingua: Compressing Prompts for Accelerated Inference of Large Language Models},
author={Huiqiang Jiang and Qianhui Wu and Chin-Yew Lin and Yuqing Yang and Lili Qiu},
booktitle={The 2023 Conference on Empirical Methods in Natural Language Processing},
year={2023},
url={https://openreview.net/forum?id=ADsEdyI32n}
}

@inproceedings{
zhang2023ho,
title={H2O: Heavy-Hitter Oracle for Efficient Generative Inference of Large Language Models},
author={Zhenyu Zhang and Ying Sheng and Tianyi Zhou and Tianlong Chen and Lianmin Zheng and Ruisi Cai and Zhao Song and Yuandong Tian and Christopher Re and Clark Barrett and Zhangyang Wang and Beidi Chen},
booktitle={Thirty-seventh Conference on Neural Information Processing Systems},
year={2023},
url={https://openreview.net/forum?id=RkRrPp7GKO}
}

@inproceedings{10.1145/3534678.3539260,
author = {Kim, Sehoon and Shen, Sheng and Thorsley, David and Gholami, Amir and Kwon, Woosuk and Hassoun, Joseph and Keutzer, Kurt},
title = {Learned Token Pruning for Transformers},
year = {2022},
isbn = {9781450393850},
publisher = {Association for Computing Machinery},
address = {New York, NY, USA},
url = {https://doi.org/10.1145/3534678.3539260},
doi = {10.1145/3534678.3539260},
booktitle = {Proceedings of the 28th ACM SIGKDD Conference on Knowledge Discovery and Data Mining},
pages = {784–794},
numpages = {11},
location = {Washington DC, USA},
series = {KDD '22}
}

@inproceedings{
li2023compressing,
title={Compressing Context to Enhance Inference Efficiency of Large Language Models},
author={Yucheng Li and Bo Dong and Frank Guerin and Chenghua Lin},
booktitle={The 2023 Conference on Empirical Methods in Natural Language Processing},
year={2023},
url={https://openreview.net/forum?id=cjbdRN8Yxy}
}

@inproceedings{lee-hockenmaier-2025-evaluating,
    title = "Evaluating Step-by-step Reasoning Traces: A Survey",
    author = "Lee, Jinu  and
      Hockenmaier, Julia",
    editor = "Christodoulopoulos, Christos  and
      Chakraborty, Tanmoy  and
      Rose, Carolyn  and
      Peng, Violet",
    booktitle = "Findings of the Association for Computational Linguistics: EMNLP 2025",
    month = nov,
    year = "2025",
    address = "Suzhou, China",
    publisher = "Association for Computational Linguistics",
    url = "https://aclanthology.org/2025.findings-emnlp.94/",
    doi = "10.18653/v1/2025.findings-emnlp.94",
    pages = "1789--1814",
    ISBN = "979-8-89176-335-7"
}

@inproceedings{jang-etal-2025-verbosity,
    title = "Verbosity-Aware Rationale Reduction: Sentence-Level Rationale Reduction for Efficient and Effective Reasoning",
    author = "Jang, Joonwon  and
      Kim, Jaehee  and
      Kweon, Wonbin  and
      Lee, Seonghyeon  and
      Yu, Hwanjo",
    editor = "Che, Wanxiang  and
      Nabende, Joyce  and
      Shutova, Ekaterina  and
      Pilehvar, Mohammad Taher",
    booktitle = "Findings of the Association for Computational Linguistics: ACL 2025",
    month = jul,
    year = "2025",
    address = "Vienna, Austria",
    publisher = "Association for Computational Linguistics",
    url = "https://aclanthology.org/2025.findings-acl.1068/",
    doi = "10.18653/v1/2025.findings-acl.1068",
    pages = "20769--20784",
    ISBN = "979-8-89176-256-5"
}

@misc{lanham2023measuring,
      title={Measuring Faithfulness in Chain-of-Thought Reasoning}, 
      author={Tamera Lanham and Anna Chen and Ansh Radhakrishnan and Benoit Steiner and Carson Denison and Danny Hernandez and Dustin Li and Esin Durmus and Evan Hubinger and Jackson Kernion and Kamile Lukosiute and Karina Nguyen and Newton Cheng and Nicholas Joseph and Nicholas Schiefer and Oliver Rausch and Robin Larson and Sam McCandlish and Sandipan Kundu and Saurav Kadavath and Shannon Yang and Thomas Henighan and Timothy Maxwell and Timothy Telleen-Lawton and Tristan Hume and Zac Hatfield-Dodds and Jared Kaplan and Jan Brauner and Samuel R. Bowman and Ethan Perez},
      year={2023},
      eprint={2307.13702},
      archivePrefix={arXiv},
      primaryClass={cs.AI},
      url={https://arxiv.org/abs/2307.13702}, 
}

@misc{xu2025chain,
      title={Chain of Draft: Thinking Faster by Writing Less}, 
      author={Silei Xu and Wenhao Xie and Lingxiao Zhao and Pengcheng He},
      year={2025},
      eprint={2502.18600},
      archivePrefix={arXiv},
      primaryClass={cs.CL},
      url={https://arxiv.org/abs/2502.18600}, 
}

@inproceedings{song-etal-2025-prmbench,
    title = "{PRMB}ench: A Fine-grained and Challenging Benchmark for Process-Level Reward Models",
    author = "Song, Mingyang  and
      Su, Zhaochen  and
      Qu, Xiaoye  and
      Zhou, Jiawei  and
      Cheng, Yu",
    editor = "Che, Wanxiang  and
      Nabende, Joyce  and
      Shutova, Ekaterina  and
      Pilehvar, Mohammad Taher",
    booktitle = "Proceedings of the 63rd Annual Meeting of the Association for Computational Linguistics (Volume 1: Long Papers)",
    month = jul,
    year = "2025",
    address = "Vienna, Austria",
    publisher = "Association for Computational Linguistics",
    url = "https://aclanthology.org/2025.acl-long.1230/",
    doi = "10.18653/v1/2025.acl-long.1230",
    pages = "25299--25346",
    ISBN = "979-8-89176-251-0"
}

\clearpage

\appendix

% ====================================================================
% 1. DATASET CONSTRUCTION (구축 알고리즘 & 통계)
% ====================================================================
\section{RIV-GSM8K Construction Details}
\label{app:construction_details}

In this section, we provide the detailed algorithm and statistical breakdown of the \BENCHMARK{} dataset construction.

\subsection{Construction Algorithm}
Algorithm~\ref{alg:riv_construction} outlines the step-by-step procedure for injecting inefficiencies into the GSM8K dataset.

% [Algorithm 1]
\begin{algorithm}[h]
\small
\caption{\small RIV-GSM8K Benchmark Construction}
\label{alg:riv_construction}
\begin{algorithmic}[1]
\REQUIRE Dataset $\mathcal{D}_{GSM8K} = \{(Q, C)\}$, Generator $\mathcal{G}$
\REQUIRE Types $\mathcal{T} = \{\text{Dup, Para, Dec, Circ, Irr}\}$
\ENSURE RIV-GSM8K Dataset $\mathcal{D}_{RIV}$

\STATE $\mathcal{D}_{RIV} \leftarrow \emptyset$; \quad $\mathcal{I}_{\mathcal{T}} \leftarrow \text{Cycle}(\mathcal{T})$

\FORALL{$(Q, C=\{S_1, \dots, S_N\}) \in \mathcal{D}_{GSM8K}$}
    \STATE $T \leftarrow \text{next}(\mathcal{I}_{\mathcal{T}})$ 
    \STATE Pick random index $t \in [1, N]$ \COMMENT{Calc. steps only if $T{=}\text{Dec}$}
    \IF{$T == \text{Simple Dup}$}
        \STATE $S_{new} \leftarrow S_t$
    \ELSE
        \STATE $S_{new} \leftarrow \mathcal{G}(Q, C, S_t \mid T)$
    \ENDIF
    \STATE \textbf{Update:} If $T{=}\text{Dec}$, replace $S_t$ with $S_{new}$; else insert $S_{new}$.
    \STATE Add $(Q, C_{new}, T)$ to $\mathcal{D}_{RIV}$
\ENDFOR
\RETURN $\mathcal{D}_{RIV}$
\end{algorithmic}
\end{algorithm}

\begin{table}[h]
\centering
\label{tab:dataset_stats}
\resizebox{\linewidth}{!}{%
\begin{tabular}{l|c|c|c}
\hline
\textbf{Perturbation Type} & \textbf{Samples} & \textbf{\# Aug. Steps} & \textbf{\# Norm. Steps} \\ \hline
Simple Duplication & 1,495 & 1,495 & 6,829 \\
Paraphrase & 1,495 & 1,495 & 6,849 \\
Decompose & 1,495 & 7,834 & 5,364 \\
Circular Reasoning & 1,494 & 3,955 & 6,857 \\
Irrelevant & 1,494 & 5,402 & 6,811 \\ \hline
\textbf{Total} & \textbf{7,473} & \textbf{20,181} & \textbf{32,780} \\ \hline
\end{tabular}%
}
\caption{Statistics of the RIV-GSM8K Benchmark. We explicitly distinguish between Augmented Steps (injected noise) and Normal Steps (baseline reasoning).}
\label{tab:dataset_stats}
\end{table}

\subsection{Dataset Statistics}
\label{app:dataset_stats}
Table~\ref{tab:dataset_stats} summarizes the distribution of the five inefficiency types within the constructed benchmark. We explicitly distinguish between 'Augmented Steps' (the injected noise) and 'Normal Steps' (the original baseline reasoning).

\section{Prompt Details and Configuration}
\label{app:prompt_details}

We utilize \texttt{gpt-4o} for all augmentation tasks. To ensure generation diversity and structural consistency, we employ a \textbf{temperature of 1.0} and a \textbf{maximum token limit of 5,000}. Additionally, we enforce a \textbf{strict JSON Schema} for all outputs to facilitate robust parsing.

\begin{figure*}[h]
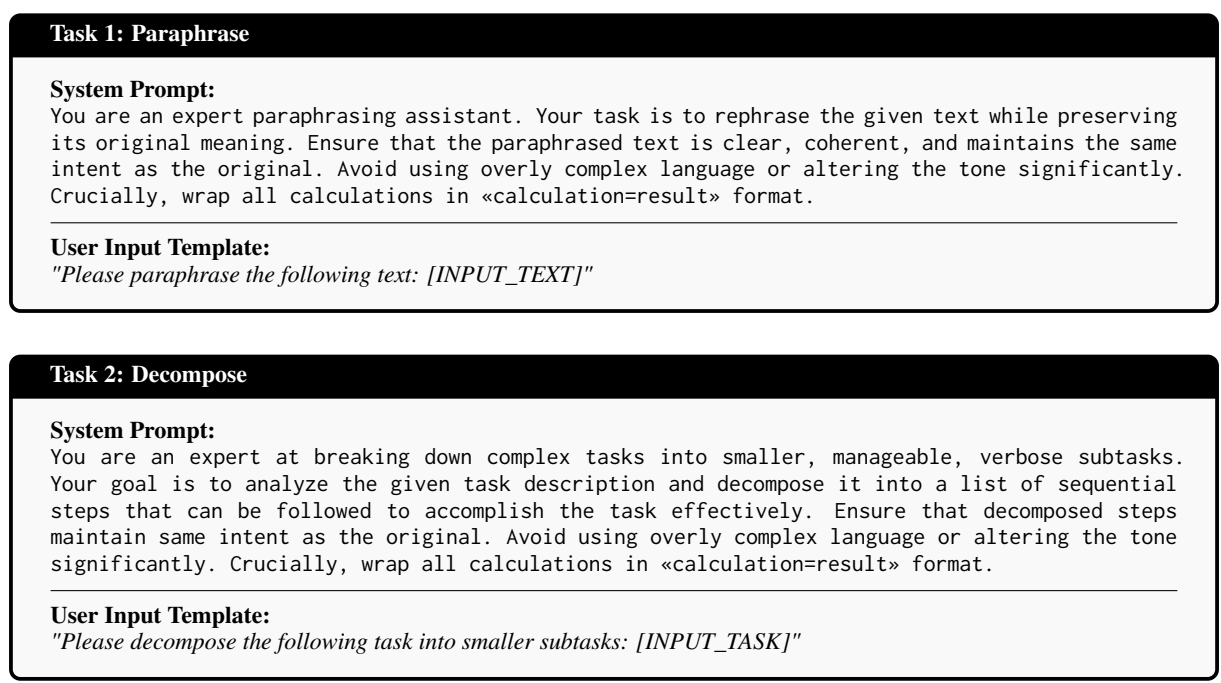
 % [t!]는 페이지 상단(Top) 고정
    \centering
    \small
    
    % [Part B] Task 1: Paraphrase
    \begin{tcolorbox}[colback=gray!5, colframe=black, title=\textbf{Task 1: Paraphrase}]
    \textbf{System Prompt:} \\
    \texttt{You are an expert paraphrasing assistant. Your task is to rephrase the given text while preserving its original meaning. Ensure that the paraphrased text is clear, coherent, and maintains the same intent as the original. Avoid using overly complex language or altering the tone significantly. Crucially, wrap all calculations in <<calculation=result>> format.}
    
    \vspace{2mm} \hrule \vspace{2mm}
    
    \textbf{User Input Template:} \\
    \textit{"Please paraphrase the following text: [INPUT\_TEXT]"}
    \end{tcolorbox}
    
    \vspace{2mm}
    
    % [Part C] Task 2: Decompose
    \begin{tcolorbox}[colback=gray!5, colframe=black, title=\textbf{Task 2: Decompose}]
    \textbf{System Prompt:} \\
    \texttt{You are an expert at breaking down complex tasks into smaller, manageable, verbose subtasks. Your goal is to analyze the given task description and decompose it into a list of sequential steps that can be followed to accomplish the task effectively. Ensure that decomposed steps maintain same intent as the original. Avoid using overly complex language or altering the tone significantly. Crucially, wrap all calculations in <<calculation=result>> format.}
    
    \vspace{2mm} \hrule \vspace{2mm}
    
    \textbf{User Input Template:} \\
    \textit{"Please decompose the following task into smaller subtasks: [INPUT\_TASK]"}
    \end{tcolorbox}
    
    \caption{Configuration and prompt details for basic augmentation strategies (\textit{Paraphrase} and \textit{Decompose}). The top block shows shared hyperparameters.}
    \label{fig:prompt_basic}
\end{figure*}

Below, we detail the specific system prompts and input templates used for our pipeline. Figure~\ref{fig:prompt_basic} outlines the basic augmentation types, while Figure~\ref{fig:prompt_context} details the context-aware strategies.

% ====================================================================
% 3. SYNTHETIC EXAMPLES (데이터 생성 결과 예시 - 프롬프트 섹션 직후 배치)
% ====================================================================
\section{Qualitative Examples of Reasoning Augmentations}
\label{app:qualitative_examples}

To better understand the nature of the perturbations introduced by our pipeline, we provide concrete examples of generated reasoning steps in Table~\ref{tab:augmentation_examples}. These examples are derived from a single original step in the GSM8K dataset: \textit{"Adults = 10 * 8 = 80"}.

As shown in Table~\ref{tab:augmentation_examples}, our augmentation strategies cover a spectrum of ``over-reasoning'' behaviors observed in Large Language Models:
\begin{itemize}
    \item \textbf{Surface-level Redundancy:} \textit{Simple Duplication} and \textit{Paraphrase} retain the exact logic of the original step but introduce lexical variations or repetitions, testing the model's robustness to verbose phrasing.
    \item \textbf{Granularity Expansion:} \textit{Decompose} breaks down a single atomic operation into a verbose chain of micro-steps (e.g., identifying variables, stating the operation, calculating, and restating the result), significantly inflating the token count without adding deductive value.
    \item \textbf{Logical Loops:} \textit{Circular Reasoning} mimics a model's tendency to ``double-check'' itself unnecessarily. It uses inverse operations (division/multiplication) to verify an already established fact, creating a closed logical loop that adds computational cost.
    \item \textbf{Contextual Noise:} \textit{Irrelevant} introduces distractors that differ from hallucinations; they are mathematically true and contextually plausible (e.g., discussing ticket prices or family groups) but contribute nothing to the solution path.
\end{itemize}

These qualitative samples illustrate the diverse challenges our dataset poses to reasoning evaluators.

\begin{table*}[h]
\centering
\small
\renewcommand{\arraystretch}{1.4} % 행 간격을 넓혀 가독성 확보

\begin{tabular}{p{0.26\textwidth} | p{0.68\textwidth}} % 오른쪽 너비를 조금 더 확보했습니다 (0.64 -> 0.68)
\toprule
\textbf{Augmentation Type} & \textbf{Generated Reasoning Step} \\
\midrule

% Original (Reference)
\rowcolor{gray!10} % 배경색을 넣어 원본임을 강조
\textbf{Original Input} & \textbf{Adults = 10 * 8 = \$<<10*8=80>>80} \\
\midrule

% 1. Simple Duplication
\textbf{1. Simple Duplication} & 
Adults = 10 * 8 = \$<<10*8=80>>80 \\
\addlinespace[1.5ex]

% 2. Paraphrase
\textbf{2. Paraphrase} & 
Adults: 10 multiplied by 8 equals \$<<10*8=80>>80 \\
\addlinespace[1.5ex]

% 3. Decompose
\textbf{3. Decompose} & 
Identify the number of adults involved, which is given as 10. $\rightarrow$ Identify the multiplication factor for the adults, which is 8 in this case. $\rightarrow$ Multiply the number of adults (10) by the factor (8): <<calculation=80>>. $\rightarrow$ State the result of the multiplication, which is \$80. \\
\addlinespace[1.5ex]

% 4. Circular Reasoning
\textbf{4. Circular Reasoning} & 
Having calculated \$80 for the adults, we can verify this by noting that \$80 divided by the 10 adults gives us \$<<80/10=8>>8 per adult. $\rightarrow$ Re-multiplying this \$8 per adult by 10 adults returns \$<<8*10=80>>80, ensuring our number is correct. $\rightarrow$ Alternatively, if we consider twice the accounted \$80, we get \$<<80*2=160>>160, and half of this sum is \$<<160/2=80>>80, which confirms our original total remains consistent at \$80. \\
\addlinespace[1.5ex]

% 5. Irrelevant
\textbf{5. Irrelevant} & 
Since the adults paid a total of \$80, this also means they chose to spend their money on tickets rather than concessions at the game, an interesting financial decision given the usual pricing. $\rightarrow$ It's notable that the sum of \$124 offers no remainder when considering integer ticket prices, suggesting an elegance in the problem structure. $\rightarrow$ The relationship between the number of adults and children indicates a typical family group could have attended this game together.
\\

\bottomrule
\end{tabular}

% 표 제목
\caption{Examples of augmented reasoning steps generated from a single original step. The generated text aims to mimic specific reasoning flaws or stylistic variations.}
\label{tab:augmentation_examples}

\end{table*}

% ====================================================================
% 4. IMPLEMENTATION DETAILS (CAID/PACE 구현 상세 - 데이터 섹션 마무리 후 시작)
% ====================================================================
\section{Implementation Details}
\label{app:implementation_details}

\subsection{Model Configuration}
\METRIC{} is designed to be computationally efficient and widely applicable. We employ the following off-the-shelf models:
\begin{itemize}
    \item \textbf{Semantic Encoder ($\mathcal{E}$):} We use \texttt{all-MiniLM-L6-v2} (22M parameters) to compute cosine similarity for Local Similarity ($\mathcal{M}_{sim}$) and Global Goal Alignment ($\mathcal{M}_{rel}$). This model was selected for its high speed and strong performance on semantic textual similarity tasks.
    \item \textbf{Density Estimator ($\mathcal{M}$):} We use \texttt{GPT-2 Small} (124M parameters) to calculate the perplexity for Information Density ($\mathcal{M}_{density}$).
\end{itemize}
The total parameter count for \METRIC{} is approximately 146M, which is significantly smaller than the baseline PRMs (e.g., ReasonEval-34B).

\subsection{Hyperparameters}
We utilize a fixed set of thresholds across all experiments (GSM8K, StrategyQA, ARC-Challenge) to demonstrate the generalizability of our metric. The specific values are:
\begin{itemize}
    \item \textbf{Removal Thresholds (\textsc{Prune}):}
    \begin{itemize}
        \item High Redundancy: $\tau_{sim} = 0.85$
        \item Low Relevance: $\tau_{rel} = 0.25$
    \end{itemize}
    \item \textbf{Compression Candidates (\textsc{Merge}):}
    \begin{itemize}
        \item Low Information Density: $\tau_{density} = 0.1$
        \item Low Semantic Delta (Base): $\tau_{delta} = 0.03$
    \end{itemize}
    \item \textbf{Adaptive Decay:}
    \begin{itemize}
        \item Decay Factor: $\lambda = 0.95$ (Applied as $\tau_{\delta}(t) = \tau_{delta} \cdot \lambda^{t}$)
    \end{itemize}
\end{itemize}

\begin{table*}[t]
\centering
\resizebox{\textwidth}{!}{%
\begin{tabular}{cl|cc|cc|c}
\toprule
\multirow{2}{*}{\textbf{ID}} & \multirow{2}{*}{\textbf{Method Description}} & \multicolumn{2}{c|}{\textbf{Performance}} & \multicolumn{2}{c|}{\textbf{Efficiency}} & \textbf{Ratio} \\
 & & Acc (\%) & $\Delta$ & Tok Red (\%) & Step Red (\%) & (Tok) \\ \midrule
0 & Baseline (Original CoT) & 82.03 & 0.00 & 0.00 & 0.00 & 1.00 \\ \midrule
1 & + Similarity (Remove) & 84.46 & +2.43 & 7.36 & 15.47 & 1.08 \\
2 & + Relevance (Remove) & 84.38 & +2.35 & 8.91 & 17.06 & 1.10 \\
3 & + Density (Remove) & \textbf{84.53} & \textbf{+2.50} & 20.12 & 21.69 & 1.25 \\
4 & + Delta (Remove) & 76.65 & -5.38 & 11.18 & \textbf{72.52} & 1.13 \\ \midrule
5 & + Merge (No Safety) & 76.95 & -5.08 & \textbf{45.09} & 70.41 & \textbf{1.82} \\
\rowcolor{gray!10} 6 & \textbf{PACE (Full Method)} & 81.12 & -0.91 & 31.05 & 53.53 & 1.45 \\
\bottomrule
\end{tabular}%
}
\caption{Ablation study of PACE components. We analyze the impact of each module on accuracy and compression efficiency. \textbf{Modes 1--4} use removal-only logic, while \textbf{Modes 5--6} introduce the merging mechanism. \textbf{PACE (Mode 6)} achieves the best balance between accuracy recovery and token reduction.}
\label{tab:PACE_ablation_tab}
\end{table*}

\paragraph{Sensitivity Analysis.}
We observed that the performance of \METRIC{} is relatively stable around these threshold values. For instance, varying $\tau_{sim}$ between 0.80 and 0.90 or $\tau_{rel}$ between 0.20 and 0.30 resulted in minimal fluctuations in the Step Preservation Rate (SPR) on the \BENCHMARK{} validation set. This suggests that the chosen hyperparameters are robust and not overfitted to a specific dataset distribution.

% ====================================================================
% 5. GOLD DATA ANALYSIS (Latent Inefficiency 분석)
% ====================================================================
\section{Qualitative Analysis of Latent Inefficiency}
\label{app:qualitative_analysis_gold}

To better understand the nature of ``Latent Inefficiency'' in human-written Gold data, we provide a detailed qualitative analysis of steps flagged for \textsc{Merge} by \METRIC{}. Table~\ref{tab:gold_merge_examples} presents concrete examples from the GSM8K dataset.

\begin{table}[h]
\centering
\small
\renewcommand{\arraystretch}{1.4} % 행간 조정
\setlength{\tabcolsep}{3pt} % 컬럼 간 여백을 조금 줄여서 헤더 공간 확보
\resizebox{\linewidth}{!}{
\begin{tabular}{
    >{\raggedright\arraybackslash}p{0.42\linewidth} % 너비를 늘리고 왼쪽 정렬
    >{\raggedright\arraybackslash}p{0.42\linewidth} % 너비를 늘리고 왼쪽 정렬
    l
}
\toprule
\textbf{Previous Step ($S_{t-1}$)} & \textbf{Target Step ($S_t$) [Gold]} & \textbf{Reason} \\ 
\midrule

% --- Example 1 ---
\textit{Child = 44/11 = \$<<44/11=4>>4} & 
\textit{Each child's ticket is \$<<4=4>>4.} & 
\textbf{\textcolor{brown}{Low Delta}} \\

% Diagnosis Row
\multicolumn{3}{>{\arraybackslash}p{\linewidth}}{
    \footnotesize \textcolor{gray}{$\rightarrow$ \textit{\textbf{Diagnosis:} The target step merely repeats the value '4' established in the previous step, contributing no new deductive information (No Progress).}}
} \\ 

\midrule

% --- Example 2 ---
\textit{...when self-checkout is broken... 160 * 1.2 = 192 complaints/day} & 
\textit{Then multiply the number of complaints per day by the number of days...: 192 * 3 = 576...} & 
\textbf{\textcolor{teal}{Low Density}} \\

% Diagnosis Row
\multicolumn{3}{>{\arraybackslash}p{\linewidth}}{
    \footnotesize \textcolor{gray}{$\rightarrow$ \textit{\textbf{Diagnosis:} The step explicitly describes the operation before performing it, inflating token usage (Verbose).}}
} \\ 

\bottomrule
\end{tabular}
}
\caption{Qualitative examples of Gold steps flagged for \textsc{Merge}. By utilizing the full width for diagnosis, we clarify why valid steps are identified as inefficient (e.g., lack of progress or verbosity).}
\label{tab:gold_merge_examples}
\end{table}

% ====================================================================
% 6. ABLATION STUDY (PACE 검증 실험)
% ====================================================================
\section{Detailed Ablation on Compression Strategy}
\label{app:pace_ablation}

In this section, we provide the extended ablation study on the compression strategies employed in \FRAMEWORK{}, validating the necessity of the \textsc{Merge} action and safety constraints.

\paragraph{Deletion vs. Compression.}
As shown in Table~\ref{tab:PACE_ablation_tab}, while removing redundant steps (Modes 1--3) yields slight accuracy gains, simply deleting steps with low logical progress (Mode 4, Delta) causes a sharp accuracy drop (-5.38\%). This implies that even repetitive or slow-progressing steps serve as essential ``connective tissue'' in the reasoning chain, carrying implicit dependencies. They cannot be blindly removed (Prune) but must be \textbf{merged} to preserve logical continuity while reducing verbosity.

\paragraph{Necessity of Safety Constraints.}
Mode 5 (Merge without constraints) achieves high token reduction (45\%) but suffers significant accuracy degradation (-5.08\%) due to semantic drift and information overload. By enforcing our safety constraints (Consistency and Saturation), \textbf{\FRAMEWORK{} (Mode 6)} successfully recovers the accuracy (Acc 81.12\%) while still delivering substantial efficiency (31\% token reduction), demonstrating that our density-aware merging strategy achieves the optimal trade-off between compression and validity.

% --- [2] minipage 제거하고 다시 figure* 사용 ---

% ============================================================
% [PART 2] Context-Aware Tasks (Full Content 반영)
% ============================================================
\begin{figure*}[h!]
\small
\centering

% Task 3: Circular Reasoning
\begin{tcolorbox}[colback=gray!5, colframe=black, title=\textbf{Task 3: Circular Reasoning}]
\textbf{System Prompt:} \\
\texttt{You are an expert at inserting circular reasoning into mathematical solutions. Your task is to generate a sequence of steps that redundantly verifies a previously established fact or calculated number using inverse operations or self-referential logic.}

\vspace{1mm}
\texttt{You will be provided with:}
\begin{enumerate}
    \setlength\itemsep{0em} \setlength\parskip{0em} \setlength\leftmargin{1.5em}
    \item \texttt{The Question}
    \item \texttt{Previous Reasoning Steps}
    \item \texttt{The Current Target Step}
\end{enumerate}

\texttt{Generate a reasoning section that:}
\begin{itemize}
    \setlength\itemsep{0em} \setlength\parskip{0em} \setlength\leftmargin{1.5em}
    \item \texttt{Takes a number or fact already established in the 'Previous Reasoning Steps'.}
    \item \texttt{Performs a set of operations that eventually lead back to the original number (e.g., "Since X is 5, multiplying by 2 gives 10, and dividing by 2 returns 5, confirming X is indeed 5.").}
    \item \texttt{Is mathematically true but strictly unnecessary for solving the problem.}
    \item \texttt{Does not alter the final answer or the logical path required for the solution.}
\end{itemize}

\texttt{Crucially, wrap all calculations in <<calculation=result>> format.}

\vspace{2mm} \hrule \vspace{2mm}

\textbf{User Input Template:} \\
\textit{"Based on the context below, generate circular reasoning sentences that could be inserted after the Current Step:"}
\end{tcolorbox}

\vspace{3mm}

% Task 4: Irrelevant (Hard)
\begin{tcolorbox}[colback=gray!5, colframe=black, title=\textbf{Task 4: Irrelevant (Hard)}]
\textbf{System Prompt:} \\
\texttt{You are an expert at generating context-aware distractions. Your task is to generate mathematically correct but irrelevant sentences that sound like they belong to the solution flow but do not advance the solution logic or provide any new information needed for the answer.}

\vspace{1mm}
\texttt{You will be provided with:}
\begin{enumerate}
    \setlength\itemsep{0em} \setlength\parskip{0em} \setlength\leftmargin{1.5em}
    \item \texttt{The Question}
    \item \texttt{Previous Reasoning Steps}
    \item \texttt{The Current Target Step}
    \item \texttt{Next Reasoning Steps}
\end{enumerate}

\texttt{Generate a reasoning section that:}
\begin{itemize}
    \setlength\itemsep{0em} \setlength\parskip{0em} \setlength\leftmargin{1.5em}
    \item \texttt{naturally fits between the previous reasoning, the current target step, and the next reasoning steps,}
    \item \texttt{maintains the same tone, context, and mathematical domain,}
    \item \texttt{uses a smooth transitional phrase to connect the surrounding steps,}
    \item \texttt{is mathematically true but does not contribute to solving the problem,}
    \item \texttt{does not alter any variables, numbers, or assumptions in the reasoning,}
    \item \texttt{and does not suggest new solution paths or constraints.}
\end{itemize}

\texttt{Crucially, wrap all calculations in <<calculation=result>> format if any numbers appear.}

\vspace{2mm} \hrule \vspace{2mm}

\textbf{User Input Template:} \\
\textit{"Based on the context below, generate irrelevant sentences that could be inserted after the Current Step:"}
\end{tcolorbox}

\caption{Prompt configurations for context-aware augmentation types (\textit{Circular Reasoning} and \textit{Irrelevant}). Note that these tasks require full context inputs (question, previous/next reasoning steps).}
\label{fig:prompt_context}
\end{figure*}

\end{document}